\documentclass{article}
\usepackage[preprint,nonatbib]{neurips_2021}
\usepackage[utf8]{inputenc} % allow utf-8 input
\usepackage[T1]{fontenc}    % use 8-bit T1 fonts
\usepackage{url}            % simple URL typesetting
\usepackage{booktabs}       % professional-quality tables
\usepackage{amsfonts}       % blackboard math symbols
\usepackage{nicefrac}       % compact symbols for 1/2, etc.
\usepackage{microtype}      % microtypography

\usepackage{url}
\usepackage{setspace}
\usepackage[dvipsnames, table]{xcolor}
\definecolor{americanrose}{rgb}{1.0, 0.01, 0.24}
\definecolor{myred}{rgb}{0.753, 0.314, 0.275}
\definecolor{myblue}{rgb}{0.0, 0.24, 0.95}
\definecolor{tbl_gray}{gray}{0.85}
\newcommand\MYhyperrefoptions{bookmarks=true,bookmarksnumbered=true,
pdfpagemode={UseOutlines},plainpages=false,pdfpagelabels=true,
colorlinks=true,linkcolor={americanrose},citecolor={myblue},urlcolor={blue}}
\usepackage[\MYhyperrefoptions,pdftex]{hyperref}

\usepackage[utf8]{inputenc}
\usepackage[small]{caption}

\usepackage{graphicx,overpic,wrapfig}

\usepackage{amsmath}
\usepackage{amsthm}
\usepackage{booktabs}
\usepackage[ruled,vlined]{algorithm2e}
\usepackage[thinlines]{easytable}
\usepackage{array,multirow,booktabs}
\usepackage{diagbox}
\usepackage{makecell}
\usepackage{arydshln}
\usepackage{multirow}
\usepackage{multicol}
\usepackage[ruled,vlined]{algorithm2e}
\usepackage{algorithmic}

\usepackage{xcolor}

\usepackage{cleveref}
\crefname{equation}{Eq.}{Eq.}
\crefname{figure}{Fig.}{Fig.}
\crefname{table}{Tab.}{Tab.}
\crefname{section}{Sec.}{Sec.}

\def\etal{\emph{et al.~}}
\def\ie{\emph{i.e.,~}}
\def\eg{\emph{e.g.,~}}

\newcommand{\mypar}[1]{\vspace{0pt}\noindent\textbf{#1}}

\newcommand{\etcite}[1]{~\etal~\cite{#1}}

\title{Adaptive Feature Alignment for Adversarial Training}

\author{%
Tao Wang$^1$
Ruixin Zhang$^1$
Xingyu Chen$^1$
Kai Zhao$^1$\thanks{Kai Zhao (kaiserzhao@tencent.com, kz@kaizhao.net) is the corresponding author.}\\
\textbf{Xiaolin Huang$^2$
Yuge Huang$^1$
Shaoxin Li$^1$
Jilin Li$^1$
Feiyue Huang$^1$}\\
\textnormal{1. Tencent Youtu Lab \ \ \ \ 2. Shanghai Jiaotong University}\\
{\tt\footnotesize
\{tobinwang,ruixinzhang,harleychen,yugehuang,darwinli,jerolinli,garyhuang\}@tencent.com}\\
{\tt\footnotesize \{xiaolinhuang\}@sjtu.edu.cn}
}

\begin{document}

\maketitle

\begin{abstract}
  Recent studies reveal that Convolutional Neural Networks (CNNs) are
  typically vulnerable to adversarial attacks, which pose a threat to security-sensitive applications.  
  Many adversarial defense methods improve robustness at the cost of accuracy,
  raising the contradiction between standard and adversarial accuracies. 
  In this paper, we observe an interesting phenomenon that
  feature statistics change monotonically and
  smoothly w.r.t the rising of attacking strength.
  Based on this observation, we propose the adaptive feature alignment (AFA) to
  generate features of arbitrary attacking strengths.
  Our method is trained to automatically align features of arbitrary
  attacking strength.
  This is done by predicting a fusing weight in a dual-BN architecture.
  Unlike previous works that need to either retrain the model or manually tune a hyper-parameters
  for different attacking strengths,
  our method can deal with arbitrary attacking strengths
  with a single model without introducing any hyper-parameter.
  Importantly, our method improves the model robustness against adversarial samples
  without incurring much loss in standard accuracy.
  Experiments on CIFAR-10, SVHN, and tiny-ImageNet datasets demonstrate that our method
  outperforms the state-of-the-art under a wide range of attacking strengths.
  The code will be made openly available upon acceptance.
\end{abstract}

\section{Introduction}
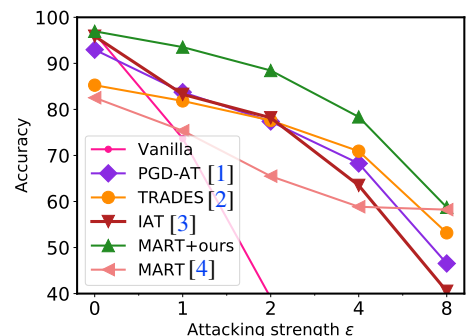
\begin{wrapfigure}{r}{0.45\textwidth}
  \vspace{-20pt}
  \begin{overpic}[width=1\linewidth]{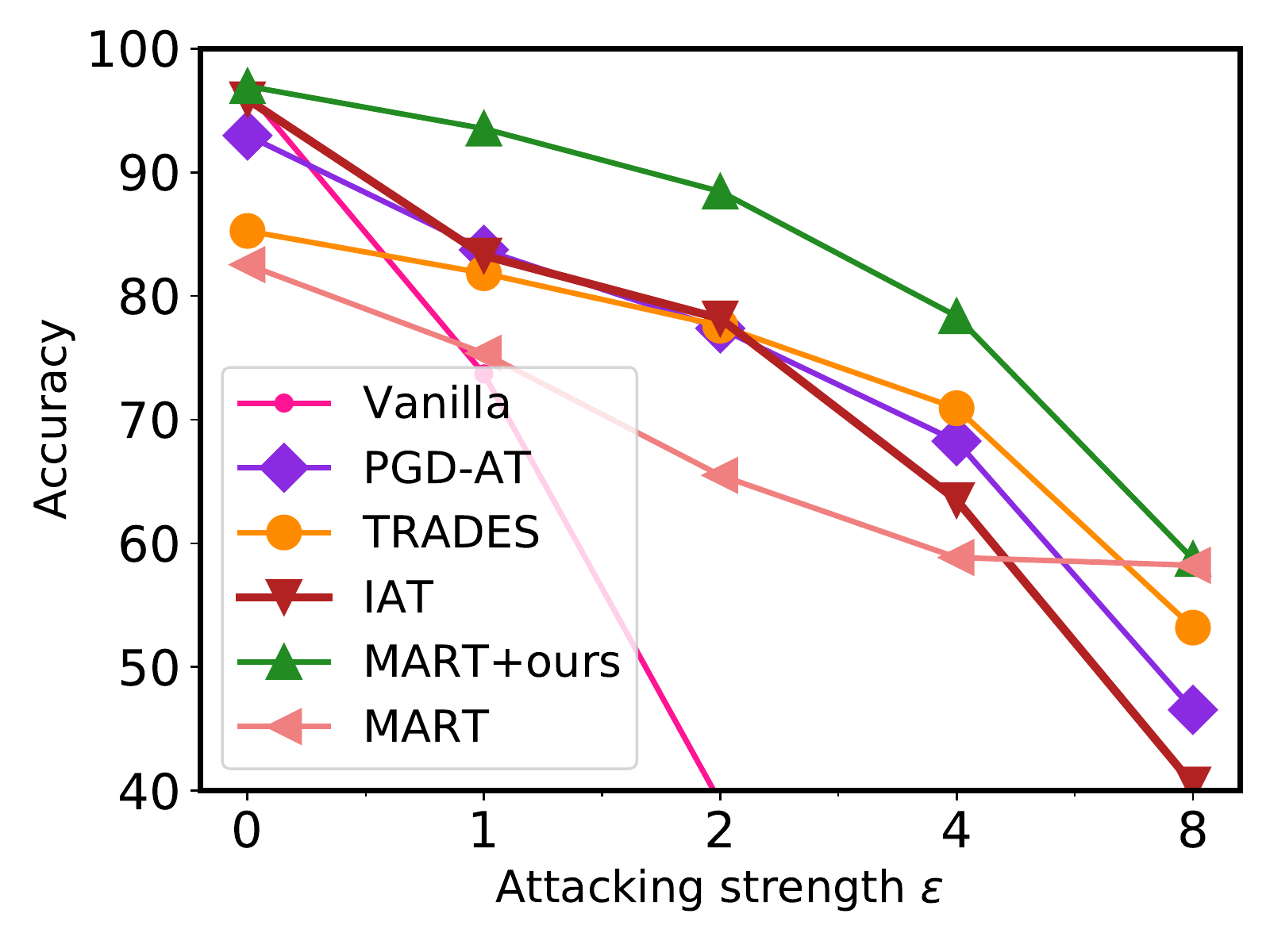}
    \put(42,36.5){\small{~\cite{madry2017towards}}} % PGD-AT
    \put(42, 31){\small{~\cite{zhang2019theoretically}}} % TRADES
    \put(34, 26){\small{~\cite{lamb2019interpolated}}} % IAT
    \put(38,16.5){\small{~\cite{wang2019improving}}} % MART
  \end{overpic}\vspace{-10pt}
  \caption{
      Performance of various methods under different attacking strength ($\epsilon$)
      on the SVHN~\cite{netzer2011reading} dataset.
      Models are implemented based on the WRN-16-8~\cite{zagoruyko2016wide} architecture.
  }\vspace{-15pt}
  \label{fig:performance_vs_eps}
\end{wrapfigure}

Recent studies reveal that convolutional neural networks (CNNs) are vulnerable to
adversarial attacks~\cite{szegedy2013intriguing,goodfellow2014explaining}, 
where human imperceptible perturbations can be crafted to fool a well-trained
network into producing incorrect predictions.
This poses a threat to security-sensitive applications, 
such as biometric identification~\cite{parkhi2015deep} and self-driving~\cite{bojarski2016end}.

A line of works have been proposed to enhance model robustness
against adversarial samples
~\cite{madry2017towards,guo2018countering,xie2018mitigating,buckman2018thermometer,xie2019feature,stutz2019disentangling}.
Among them the adversarial training (AT) methods
can achieve strong performance under a variety of attacking
configurations~\cite{madry2017towards,zhang2019theoretically}.
Though its robustness against adversarial samples,
AT-based methods
notoriously sacrifice accuracy on normal samples (standard accuracy)
~\cite{tsipras2019robustness,xie2019feature}.
As shown in ~\cref{fig:performance_vs_eps}, existing adversarial training methods
improve adversarial accuracy ($\epsilon > 0$) with the cost of standard accuracy
($\epsilon = 0$).

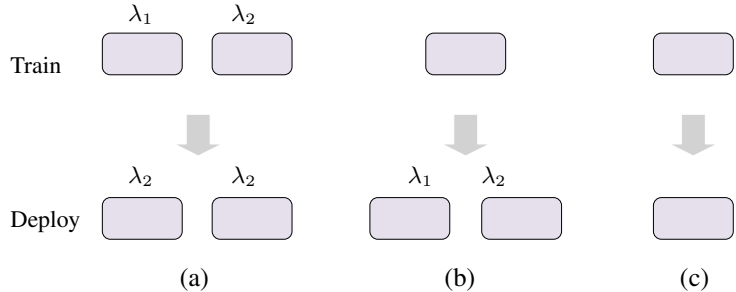
\begin{figure}[!tb]
  \centering
  \begin{overpic}[width=0.7\linewidth]{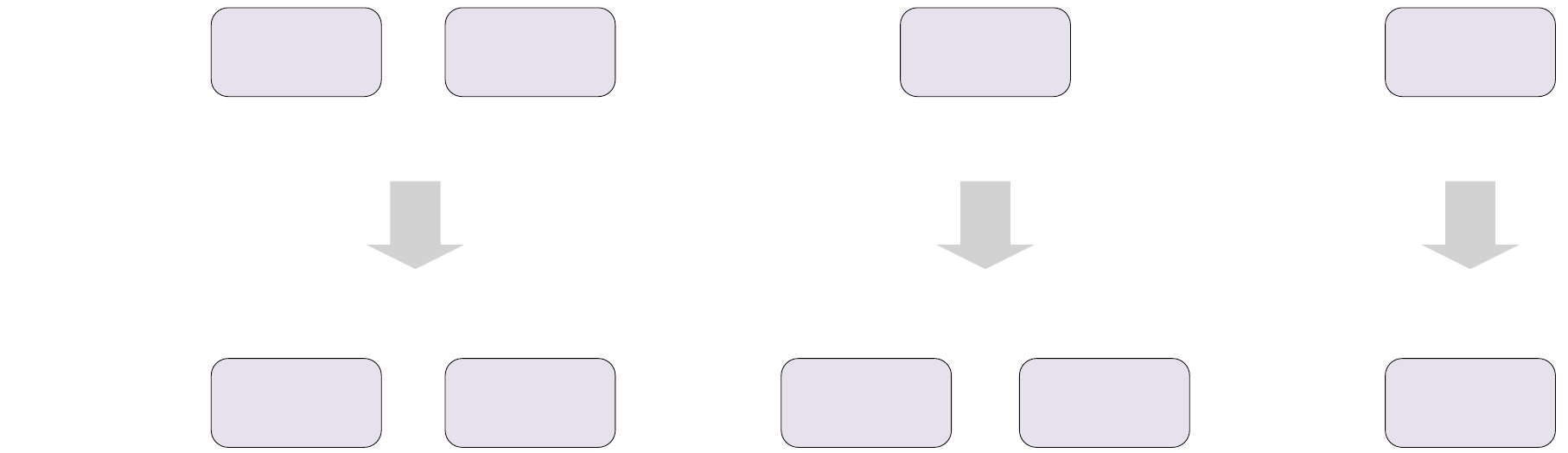}
      \put(1,23){\small{Train}}
      \put(1,2){\small{Deploy}}
      \put(24,-6){(a)}
      \put(60,-6){(b)}
      \put(92,-6){(c)}
      \put(17,30){\small{$\lambda_1$}}
      \put(31,30){\small{$\lambda_2$}}
      \put(17,8){\small{$\lambda_2$}}
      \put(31,8){\small{$\lambda_2$}}
      \put(55,8){\small{$\lambda_1$}}
      \put(65,8){\small{$\lambda_2$}}
  \end{overpic}\vspace{22pt}
  \caption{
      Schemas of different methods to deal with various attacking strengths.
      (a) TRADES \protect \cite{zhang2019theoretically} trains
      multiple networks with different hyper-parameters, \eg $\lambda$.
      (b) OAT \protect \cite{wang2020onceforall} trains only one network
      and then deploy it with various hyper-parameters.
      (c) Our proposed method needs only one network in both training
      and deployment.
  }
  \label{fig:intro_framework}
\end{figure}

The contradiction between standard and adversarial accuracies
encourages researchers to develop techniques that can
handle various attacking strengths~\cite{zhang2019theoretically,wang2020onceforall}.
Zhang~\etcite{zhang2019theoretically} introduce a hyper-parameter $\lambda$
to control the attacking strength during training.
A model trained with different $\lambda$ can be deployed in various
environments with different attacking strengths.
However, this method needs to retrain the model whenever deploying to a new environment.
Wang~\etcite{wang2020onceforall} then propose to embed the parameter $\lambda$ as
an input of the model.
During testing, the attacking strength is fed into the model as part of the input.
consequently, mode trained only once can deal with various attacking strengths during testing.
However, in real-world applications, the attacking strength is unknown to the model.
After training only once, our proposed method can deal with samples of various
attacking strengths without introducing any hyper-parameters.
~\cref{fig:intro_framework} outlines the schematic difference between
~\cite{zhang2019theoretically}, ~\cite{wang2020onceforall} and our proposed method.

As pointed by previous research~\cite{ilyas2019adversarial,xie2020adversarial,wang2020onceforall}, 
the contradiction between standard and
adversarial accuracy may be caused by the misaligned statistics between
standard and adversarial features.
We conducted an in-depth study on adversarial samples of various attacking strengths
and found that the feature statistics undergo a smooth and monotonical transfer
with the rising of attacking strength.
In other words, features of various attacking strengths can be regarded as
a continuous domain transfer~\cite{wang2020continuously}.
This phenomenon hints us that features of an arbitrary attacking strength
can be approximated through linear interpolation of several basic attacking
strengths, we will illustrate this part in~\cref{sec:motivation}.

Inspired by our observation, we develop the novel adaptive feature alignment (AFA)
framework which automatically aligns features of arbitrary attacking strengths.
The experiments on CIFAR-10, SVHN
and tiny-imagenet~\cite{deng2009imagenet,tinyimagenet}
datasets demonstrate that our proposed method surpasses
the prior methods under various attacking strengths.
The contribution of this paper is summarized as below:
\vspace{-8pt}
\begin{enumerate}\setlength\itemsep{-0.1em}
  \item We observe that feature statistics of various attacking strengths
        undergo \emph{smooth} and \emph{monotonical} transfer.
  \item Based on the observation, we are motivated to propose the 
        adaptive feature alignment (AFA) framework that automatically
        align features for arbitrary attacking strengths by predicting
        fusing weights in a dual-BN architecture.
  \item Extensive experiments on SVHN, CIFAR-10, and tiny-ImageNet datasets
        demonstrate that our method outperforms baseline adversarial training
        methods under a wide range of attacking strengths.
\end{enumerate}

%-------------------------------------------------------------------------%
%-------------------------------------------------------------------------%
\section{Related Work}\label{sec:related}
%-------------------------------------------------------------------------
\subsection{Adversarial Attacking}
Adversarial attacking aims to applying human-imperceptible perturbations to the
original inputs and consequently mislead the deep neural network to produce
incorrect predictions.
These generated samples are commonly referred to as `adversarial samples'.
%
% Since deep neural networks are trained with the gradient back-propagation algorithm,
% a line of attacking methods utilize `gradient-ascent' iterations to generate adversarial samples.
%
Goodfellow~\etal propose the fast gradient sign method (FGSM)~\cite{goodfellow2014explaining}
which calculates perturbation to input data in a single step using the gradient of the loss
(cost) function w.r.t the input.
Iterative FGSM (I-FGSM) ~\cite{kurakin2017adversarial} iteratively performs
the FGSM attack.
In each iteration, only a fraction of the allowed noise limit is added to the inputs.
The projected gradient descent (PGD) attack is similar to ~\cite{madry2017towards} I-FGSM and the only difference is that
PGD initializes the perturbation with a random noise while I-FGSM starts from zeros.
Another popular attacking method is the Carlini and Wagner method ~\cite{carlini2017towards}
(C\&W attack).
C\&W attack investigates multiple loss functions and finds a loss function that maximizes
the gap between the target logit and the highest 
C\&W attack and PGD (Projected Gradient Descent) attack~\cite{madry2017towards}.

\subsection{Adversarial Training}
Adversarial training is an effective way of improving model robustness against adversarial
attacks.
It enhances adversarial robustness by training the model with online-generated adversarial samples.
Goodfellow \etal ~\cite{goodfellow2014explaining} use
FGSM attacked samples as training data to improve the model's robustness.
Kurakin~\etal~\cite{kurakin2017adversarial} propose the iterative FGSM to further improve the performance.
Tram{\`e}r \etal~\cite{tramer2018ensemble} propose an ensemble adversarial training with adversarial examples generated
from several pre-trained models.
Zhang \etal~\cite{zhang2019theoretically} propose TRADES that balances the
adversarial accuracy with standard accuracy.
Several improvements of PGD adversarial training have also been proposed,
such as ~\cite{yan2018deep,cisse2017parseval} and ~\cite{farnia2018generalizable}.
Interestingly, Xie~\etal~\cite{xie2020adversarial} find that adversarial samples can also improve
model perform on normal data.
Despite much attention has paid to adversarial training and substantial improvements have
been achieved, these methods still suffer from the trade-off between adversarial accuracy
and standard accuracy,
and can not improve adversarial accuracy without incurring much loss in
standard accuracy.

Zhang~\etal ~\cite{zhang2019theoretically} use a hyper-parameter to control the attacking
strength to adapt to various environments.
Consequently, a model can deal with different attacking strengths by retraining
the model with another hyper-parameter.
Wang~\etal~\cite{wang2020onceforall} then take the attacking strength as part of the input
and the model is adaptable to different attacking strengths without retraining.
However, both of the two methods cannot automatically adjust the attacking strength,
which limits their application.

\begin{figure*}[!tb]
  \includegraphics[width=\linewidth]{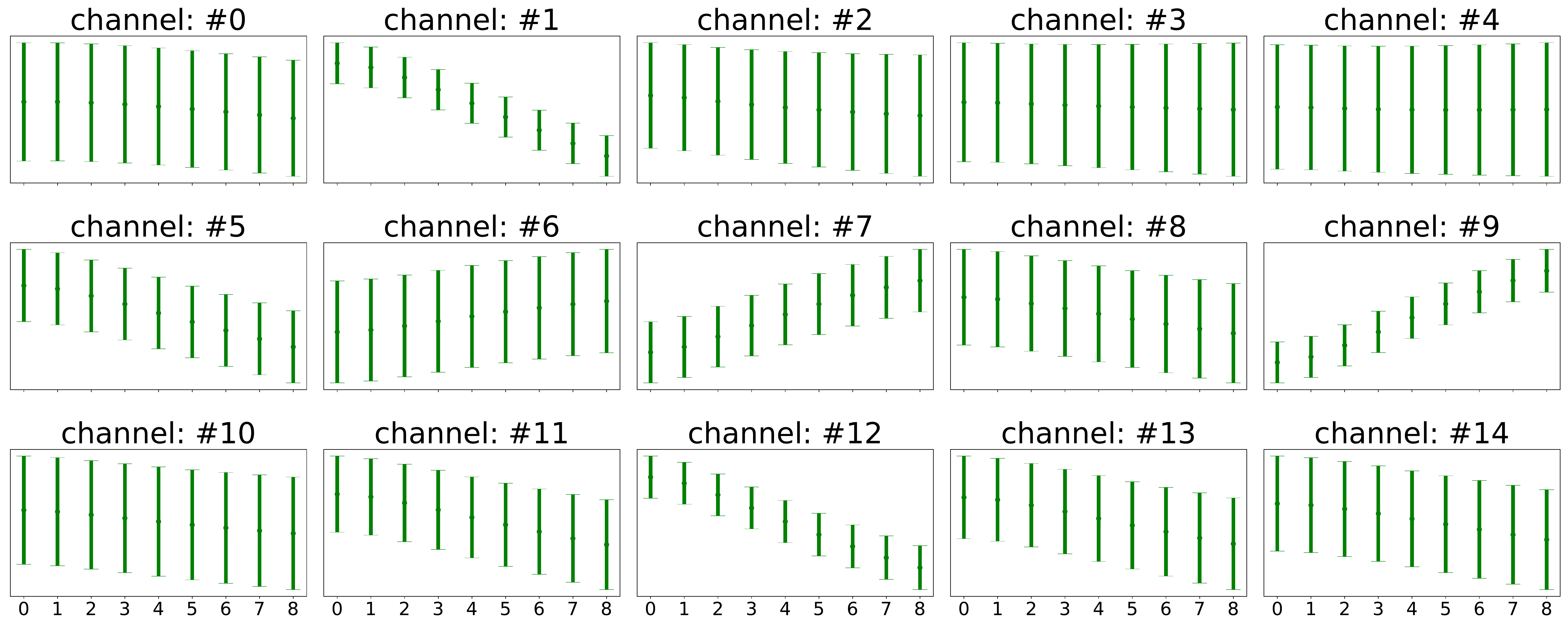}
  \vspace{-10pt}
  \caption{
      Per-channel mean and variance of intermediate convolution features under
      different attacking strengths.
      Features under different attacking strengths undergo a
      \emph{continuous} domain transfer.
  }
  \label{fig:feature-stat}
\end{figure*}
%-------------------------------------------------------------------------%
%-------------------------------------------------------------------------%
\section{Observation and Motivation}\label{sec:motivation}
In this section, we first describe the observation and
illustrate how we are motivated to develop the Adaptive Feature Alignment method.
Then we introduce the details of our proposed method.

\subsection{Observation and Motivation}\label{sec:observation}
Many recent studies reveal that features of standard/adversarial samples belong to
two separate domains
~\cite{xie2020adversarial,wang2020onceforall},
and then developed the dual-BN architecture to process standard/adversarial
samples separately.
While their methods~\cite{xie2020adversarial,wang2020onceforall} neglect the
inherent variances across adversarial samples of different attacking strengths.
Our observational experiments demonstrate that feature statistics
of variance attacking strengths undergo a \emph{continuous} and \emph{monotonical}
domain transfer.

In the observational experiment,
we train the Wide-Resnet-28~\cite{zagoruyko2016wide} model on the
CIFAR-10~\cite{krizhevsky2009learning} dataset.
We adopt the dual-BN architecture proposed in~\cite{xie2020adversarial}
which splits normal and adversarial samples into multiple
parallel batch-norm branches.
After training, we analyze the statistics of features of various attacking strength
on the testing set.
Concretely, we visualize the per-channel mean and variance of features from an intermediate
convolution layer and results are presented in ~\cref{fig:feature-stat}.

The results in ~\cref{fig:feature-stat} clearly demonstrate
that the features are transferring \emph{smoothly} and \emph{monotonically}
with the rising of attacking strength.
This hints us that features of an arbitrary attacking strength can be approximated
by a linear combination few "base" intensities.
For instance, a medium attacking strength
can be represented by a combination of a slight attacking and a strong attacking.

Based on the observational experiment, our motivations are two folds:
\begin{enumerate}\setlength\itemsep{-0.1em}
  \item Features of different attack strengths belong to their respect domains,
        and use a multi-BN architecture where each BN branch deals with a single
        attacking strength may result in better aligned features.
  \item Features of an arbitrary attacking strength can be represented by a interpolation
        of two `base attacking strengths`.
\end{enumerate}
Based on above motivation, we design a two-stage framework
to train a model that can adaptively align feature for an arbitrary
unknown attacking strength.
Next, we will detailedly describe the proposed method.

%-------------------------------------------------------------------------%
%-------------------------------------------------------------------------%
\section{Methodology}\label{sec:methodology}

\subsection{Overall Framework}
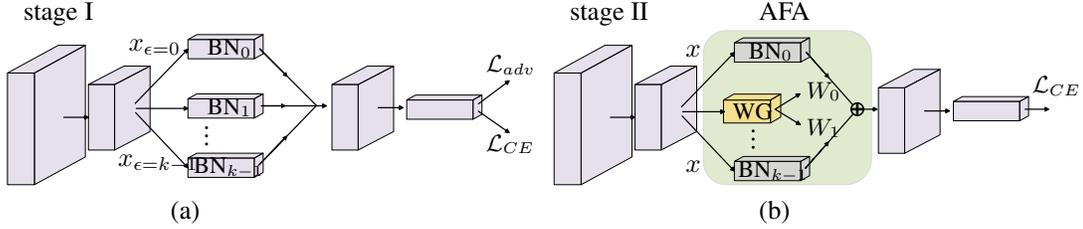
\begin{figure*}[!tb]
  \centering
  \begin{overpic}[width=1\linewidth]{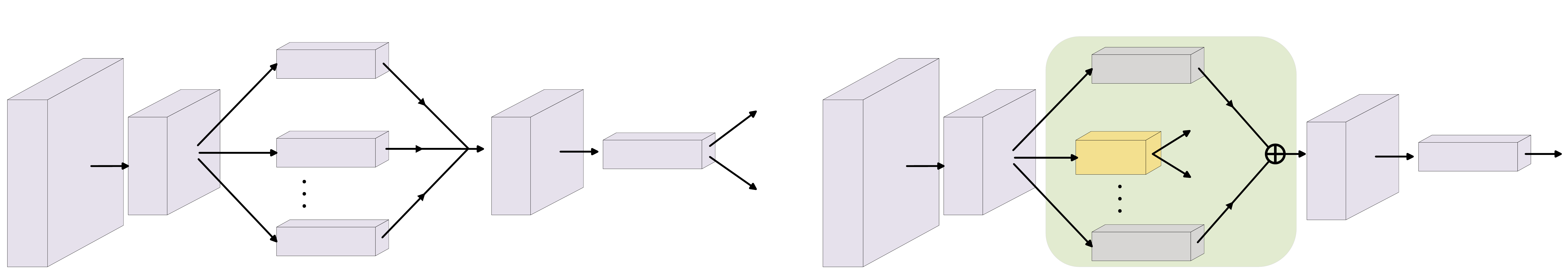}
      \put(19.5,12.6){\small{BN$_0$}}
      \put(19.5,7){\small{BN$_1$}}
      \put(18.2,1.4){\small{BN$_{k-1}$}}
      \put(12,13){$x_{\epsilon=0}$}
      % \put(13.6,8.3){$x_{\epsilon=1}$}
      \put(11,2){$x_{\epsilon=k-1}$}
      \put(46,11){\footnotesize{$\mathcal{L}_{adv}$}}
      \put(46,3.5){\footnotesize{$\mathcal{L}_{CE}$}}
      \put(65,12.5){$x$}
      \put(65,1.5){$x$}
      \put(71,12.3){\footnotesize{BN$_0$}}
      \put(70,1.0){\footnotesize{BN$_{k-1}$}}
      \put(2,16){\normalsize{stage I}}
      \put(54,16){\normalsize{stage II}}
      \put(72,16){AFA}
      \put(16,-3){(a)}
      \put(72,-3){(b)}
      \put(76.5,8.5){\footnotesize{$W_0$}}
      \put(76.5,5){\footnotesize{$W_1$}}
      \put(98,9){\footnotesize{$\mathcal{L}_{CE}$}}
      \put(69.5,6.5){\small{WG}}
  \end{overpic}\vspace{13pt}
  \caption{
      The overall architecture.
      (a) In the first stage, we train a dual-branch network where
      each parallel BN branch corresponds to samples of a specific
      attacking strength.
      (b) In the second stage, we drop intermediate BN branches
      and only keep the outermost branches,
      \ie BN$_0$ and BN$_{k-1}$.
      And the weight generator (WG) in the adaptive feature alignment (AFA)
      module will generate the fusing weight
      of the two remaining branches.
  }\label{fig:overall-architecture}
\end{figure*}
The overall architecture of our proposed framework is illustrated
in~\cref{fig:overall-architecture}.
We denote $I_{\epsilon=k}$ as an adversarial image where $\epsilon$
is the attacking strength, and $I$ represents an image of arbitrary
unknown attacking strength.
Commonly these attacking methods apply a gradient ascent to the original
inputs to obtain adversarial samples.
The attacking strength $\epsilon$ is closely related to the gradient magnitude.
Specifically, we refer to $I_{\epsilon=0}$ as normal samples without adversarial attacking.
Let $x_{\epsilon=k}$ be the convolutional feature
extracted from an intermediate layer.
Our overall pipeline is extended from the dual-BN architecture that is
proposed in~\cite{xie2020adversarial,wang2020onceforall}.
We first extend it into a \ie $K$-BN ($K\ge 2$) architecture (~\cref{fig:overall-architecture} (a)),
and then drop some of the BN branches to form the final dual-BN architecture (~\cref{fig:overall-architecture} (b)).
The training procedure of our framework includes two stages.

In the first stage (Stage I), we train a $K$-BN network with $K$ parallel batch normalization branches
and each branch BN$_k (k=0,1,...,K-1)$ deals with samples of specific attacking strength.
%
% As mentioned in ~\cref{sec:motivation}, samples of different attacking strengths
% belongs to their respect domain.
% Therefore, training a $K$-BN architecture where each BN corresponds to
% a single attacking strength leads to well aligned features.
%
Note that the attacking strength $\epsilon$ is known to the model so that each sample $x_{\epsilon}$
will only go through the corresponding BN branch according to  $\epsilon$.
~\cref{fig:overall-architecture} (a) demonstrates the training of stage I.

In the second stage (stage II), we drop the intermediate BN branches and only keep
two branches corresponding to the
strongest and lowest attacking strengths, \ie BN$_0$ and \ie BN$_{K-1}$.
Then we freeze all the model parameters and train our proposed Adaptive Feature Alignment (AFA)
module to automatically adjust fusing weight between BN$_0$ and BN$_{K-1}$.
%
% The motivation here is that, as mentioned in ~\cref{sec:motivation}, the features
% of different $\epsilon$ undergo \emph{continuous} and \emph{monotonical}
% transfer.
%
Therefore, the feature of an arbitrary $\epsilon$ can be represented by interpolation
of tw `base' attacking strengths.
~\cref{fig:overall-architecture} (b) illustrates the network structure in stage II.
Next, we will detail the proposed `adaptive feature alignment' module.

\subsection{Adaptive Feature Alignment}\label{sec:adaptive-feature-align}
\begin{figure}[!htb]
  \centering
   \begin{overpic}[width=0.75\linewidth]{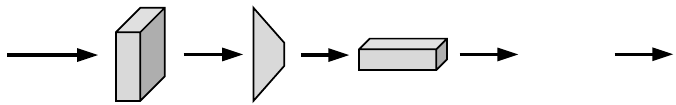}
      \put(77,6.5){\small{Sigmoid}}
      \put(10,-3){\scriptsize{Conv+BN+ReLU}}
      \put(25,12){$\times 2$}
      \put(33,-3){\scriptsize{Avg Pooling}}
      \put(56,1){\scriptsize{Linear}}
      \put(66,12){$\times 2$}
   \end{overpic}
   \vspace{15pt}
   \caption{
    The architecture of the weight generator (WG).
    The cube represents the
    `Convolution $\rightarrow$ batchnorm $\rightarrow$ ReLU'
    sequantial,
    the trapezoid represents the `global average pooling'
    and the bar represents the `Linear $\rightarrow$ ReLU'
    combo.
    The output is normalized by the sigmoid function.
   }
   \label{fig:WG-arch} 
\end{figure}
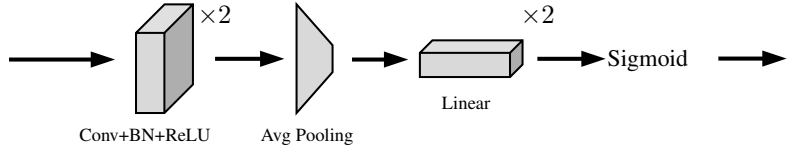

Let $x$ be the image feature of an arbitrary sample,
and BN$_0$ and BN$_{k-1}$ be the remaining BN branches as shown
in~\cref{fig:overall-architecture} (b).

Given the input $x$, the proposed AFA first generates two fusing weights $W_0, W_1$
with a subnetwork termed `weight generator' (WG).
The weight generator network consists of
2 sequential Conv-BN-ReLU blocks and then followed by the global average pooling
(AVG) and two linear layers.
Finally, the output is normalized by the sigmoid function.
~\cref{fig:WG-arch} detailly illustrates the architecture of the WG subnetwork.
consequently, the fusing weights are output of the weight generator:
\begin{equation}
  \begin{split}
  W_0 &= \text{WG}(x) \\
  W_1 &= 1 - W_0.
  \end{split}
\end{equation}

Given input $x$ and fusing weights $W_0, W_1$,
the out put of AFA $\hat{x}$ is:
\begin{equation}
  \hat{x} = W_0\cdot \text{BN}_0(x) + W_1\cdot \text{BN}_{k-1}(x),
  \label{eq:afa}
\end{equation}
where BN$_0$ and BN$_{k-1}$ are the batchnorm operator:
$$
\text{BN}(x) = \frac{x - \mu}{\sigma}.
$$

\subsection{Model Optimization}\label{subsec:pipeline}
As aforementioned, the proposed framework contains two training stages.
In the first stage, we train a $K$-BN network with both classification
loss and the adversarial loss.
We use the cross-entropy loss (CE-loss) as the classification loss.
Note that our method does not constrain the specific form of
adversarial loss.
In the second stage, we drop all middle BN-branches and only preserve two outermost
ones.
Then we fix all network parameters and only
optimize the \textbf{weight generator}, as illustrated in~\cref{fig:overall-architecture} (b).
Since the attacking strength is unknown to the
model, we use only the classification loss in stage II.

\mypar{Stage 1: Training basic model.}
In the first stage, we train the $K$-BN architecture as illustrated
in~\cref{fig:overall-architecture} (a).
Note that each BN branch only accepts samples of a specific attacking
strength and all other parameters are shared across all samples.
Given a sample $x$ with attacking strength $\epsilon$, the loss function
can be formulated as:
\begin{equation}
  \mathcal{L}_1(x_{\epsilon}) = \mathcal{L}_{CE} + 1(\epsilon>0)\cdot\mathcal{L}_{adv}
  \label{eq:overall-loss}
\end{equation}
where $1(\cdot)$ is a indicator function evaluating to 1 when the condition
is true and 0 elsewise.
The optimization procedure of stage I is summarized in
~\cref{alg:training_basic_model}, which is extended from
~\cite{xie2020adversarial}.

\mypar{Stage 2: Training the weight generator.} 
In the second stage, as illustrated in~\cref{fig:overall-architecture} (b),
we drop all other BN branches and only preserve the two outermost
branches: BN$_0$ and BN-$_{k-1}$.

Then we place the weight generator (WG)
to the end of an intermediate layer to generate adaptive fusing
weights $W$.
In stage II, we use samples of various attacking strengths to train the WG
but the attacking strength is unknown
to the model.
The only supervision in the second stage is the classification loss, as illustrated
in~\cref{fig:overall-architecture} (b).

\begin{algorithm}[!tb]
  \label{alg:stage1}
  \small
  \setstretch{1}
  \SetAlgoLined
  \KwIn{The batch of natural samples $x_c$ with label $y$, the path quantity $K$, 
  the attack strength vector of each path $\xi$, the adversarial samples generating algorithm $G$, loss function for
  natural training $L_c$, loss function for adversarial training $L_a$, the network parameters $\Theta$}
  iteration  number $i\leftarrow 0$, 
  total loss $L_i \leftarrow 0$\;
  \While{not converged}{
  $L_i \leftarrow 0$\;
  \For{$k\gets 2$ \KwTo $K$}{
    Switch to $k$th path to enable $k$th BN\;
    Obtain corresponding attack strength value from the vector $\xi_k \leftarrow \xi[k]$;
  
    Generate adversarial samples with attacking strength $\xi_k$ by $x_a \leftarrow G(\Theta, x_c, y, \xi_k)$;
  
    Calculate the adversarial loss of $k$th path by $l_k \leftarrow L_a(\Theta,x_c,x_a,y)$;
  
    Accumulate the total loss $L_i \leftarrow L_i + l_k$;
  }
  Calculate the normal loss $l_1 \leftarrow L_c(\Theta, x_c, y)$\;
  Accumulate the total loss $L_i \leftarrow L_i + l_1$\;
  Compute the gradients of $\Theta$\;
  Update the parameters $\Theta$\;
  }
  \KwOut{$\Theta$}
  \caption{Training procedure of basic model}
  \label{alg:training_basic_model}
\end{algorithm}

\section{Experiments}\label{sec:exp}

%In this section we first describe our framework in detail,
%and then report quantitative results compared with competitors.
%At last, we ablate and analyze the proposed method.
\subsection{Experiments Setup} \label{set:impl-details}
\mypar{Implementation details.}
We implement our method with the PyTorch~\cite{paszke2019pytorch} framework.
We conduct experiments on three datasets:
CIFAR-10~\cite{krizhevsky2009learning},
SVHN~\cite{netzer2011reading}
and tiny-ImageNet~\cite{deng2009imagenet,tinyimagenet}. 
Following several previous works~\cite{ding2019mma,wang2019bilateral,wang2020onceforall},
we use the WRN-28-10~\cite{zagoruyko2016wide} network as the backbone
for the experiments on CIFAR-10.
The WRN-16-8 network is used on SVHN and tiny-ImageNet datasets.
We use $K=5$ in all our experiments since we found that $K=5$ achieves
a good balance between training efficiency and model performance,
as illustrated in our ablation study in~\cref{sec:ablation}.

As aforementioned, the training procedure is separated into
two stages.
On SVHN, the initial learning rates of the two
stages are 0.01 and 0.001, respectively.
And the initial learning rates on CIFAR and tiny-ImageNet
datasets are 0.1 and 0.01, respectively.
We train 100 epochs for the first stage and 20 epochs for the second
stage.
During training, we decay the learning rate with the factor of $0.9$.
In the first stage, the learning rate decays at the 75th and the
90th epoch;
in the second stage, the learning rate decays at the 10th epoch.

\mypar{Attacking methods.}
We adopt 3 adversarial attack algorithms:
Fast Gradient Sign Method (FGSM)~\cite{goodfellow2014explaining},
Projected Gradient Descent (PGD)~\cite{madry2017towards}
and C\&W attack~\cite{carlini2017towards} .

In~\cref{tab:svhn-cifar10} and ~\cref{tab:tiny-imagenet}, we compare our
method with other adversarial training methods under the the PGD attack
with various attacking strengths, \eg $\epsilon=0, 1, 2, 4, 8$.
In ~\cref{tab:other_attack}
we test the proposed method under three attacking methods, \eg FGSM, PGD, and C\&W,
under a fixed attacking strength of $\epsilon=8$.

\mypar{Adversarial training baselines.}
As demonstrated in~\cref{eq:overall-loss}, our method does not constrain 
the specific form of the adversarial loss function.
Therefore, our method can be integrated with many adversarial training methods.
In our experiments, we use 4 adversarial training methods as baselines:
PGD-AT~\cite{madry2017towards},
TRADES~\cite{zhang2019theoretically},
IAT~\cite{lamb2019interpolated}
and MART~\cite{wang2019improving}.

\subsection{Quantitative Comparisons}\label{sec:quantitative}
We compare our proposed method with several recent methods,
\ie OAT~\cite{wang2020onceforall},PGD-AT~\cite{madry2017towards},
TRADES~\cite{zhang2019theoretically},
IAT~\cite{lamb2019interpolated},
and MART~\cite{wang2019improving}.
Quantitative results on
CIFAR-10~\cite{krizhevsky2009learning},
SVHN and tiny-ImageNet~\cite{tinyimagenet} datasets
are summarized in~\cref{tab:svhn-cifar10}
and~\cref{tab:tiny-imagenet}, respectively.

\begin{table}[!htb]
  \centering
  \resizebox{0.95\linewidth}{!}{
  \begin{tabular}{c|cccccc|cccccc}
    \toprule
    \multirow{2}{*}{Method} & \multicolumn{6}{c}{SVHN} & \multicolumn{6}{c}{CIFAR-10}\\
    & $\epsilon=0$ & $\epsilon=1$ & $\epsilon=2$ & $\epsilon=4$ & $\epsilon=8$ & Avg
    & $\epsilon=0$ & $\epsilon=1$ & $\epsilon=2$ & $\epsilon=4$ & $\epsilon=8$ & Avg \\
        \hline
    Standard & 96.5& 73.7& 39.1& 6.5 & 0.2 & 43.2 & 95.2& 30.5 & 3.0 & 0.0 & 0.0 & 25.7 \\
    \cdashline{1-7}[1pt/2pt]
    PGD-AT  & 93.0& 83.7& 77.4& 68.3& \textbf{46.5}& 73.8 & 86.9& 81.5& 78.5& 68.8& 46.2& 72.4\\
    \makecell{\textbf{PGD-AT}+ \\ \textbf{ours}} & \textbf{98.1} & \textbf{93.1} & \textbf{87.8} & \textbf{75.4} & 46.4 & \textbf{80.0} & \textbf{95.8} & \textbf{87.0} & \textbf{83.9} & \textbf{72.6} & \textbf{50.0} & \textbf{77.9}\\
    \cdashline{1-13}[1pt/2pt]
    TRADES  & 85.3& 81.8& 77.6& 70.9& 53.2& 73.8  & 83.2& 80.5& 77.6& 71.3& \textbf{53.8}& 73.3\\
    \makecell{\textbf{TRADES}+\\ \textbf{ours}}     & \textbf{97.0} & \textbf{89.0} & \textbf{88.4} & \textbf{77.6} & \textbf{54.2} & \textbf{81.4} & \textbf{95.4} & \textbf{86.8} & \textbf{82.7} & \textbf{72.1} & 52.6 & \textbf{77.9}\\
    \cdashline{1-13}[1pt/2pt]
    IAT  & 95.9& 83.3& 78.2& 63.5& 40.5& 72.3  & 92.9& \textbf{88.3}& \textbf{84.4}& 73.5& 46.2& \textbf{77.0}        \\
    \textbf{IAT+ours}  & \textbf{97.4} & \textbf{91.6} & \textbf{87.9} & \textbf{75.9} & \textbf{42.3} & \textbf{79.0} & \textbf{96.1} & 86.2 & 81.7 & \textbf{73.7} & \textbf{46.2} & 76.7  \\
    \cdashline{1-13}[1pt/2pt]
    MART & 82.5& 75.4& 65.5& 58.9& 58.2& 68.1 &  83.6& 81.0& 78.2& 72.3& 55.3& 74.1       \\
    \makecell{\textbf{MART}+\\ \textbf{ours}} & \textbf{97.0} & \textbf{93.5} & \textbf{88.5} & \textbf{78.4} & \textbf{58.7} & \textbf{83.2} & \textbf{95.9} & \textbf{84.3} & \textbf{84.2} & \textbf{73.5} & \textbf{56.1} & \textbf{78.8}\\
    \bottomrule
  \end{tabular}
  }
  \vspace{10pt}
  \caption{
    Adversarial accuracy on SVHN and CIFAR-10 datasets.
    Our proposed method consistently outperforms baseline adversarial training methods with a clear margin.
  }\label{tab:svhn-cifar10}
\end{table}

\mypar{Performance on SVHN and CIFAR-10.}
We first test the proposed method as well as other
adversarial training methods on two small datasets:
SVHN and CIFAR-10.
Results in ~\cref{tab:svhn-cifar10} reveal that:
1) our method outperforms the standard model in terms of standard accuracy ($\epsilon=0$),
this is mainly due to the dual-BN architecture~\cite{xie2020adversarial}.
2) Our proposed method consistently improve the adversarial accuracy based on three adversarial
training methods.
Especially, our method achieves high adversarial accuracy under a variety of attacking strengths,
we believe this this mainly due to the proposed `adaptive feature align' that can automatically
align features according to the attacking strength.
\begin{table}[!htb]
  \centering
  \resizebox{0.6\linewidth}{!}{
  \begin{tabular}{ccccccc}
  \toprule
  \multirow{2}{*}{Method} & \multicolumn{6}{c}{Acc under different $\epsilon$}                                                  \\ \cline{2-7} 
     & 0   & 1   & 2   & 4   & 8   & Avg           \\ \hline
  Standard      & 63.5& 6.7& 1.0& 0.2 & 0.0 & 14.3          \\ \cdashline{1-7}[1pt/2pt]
  PGD-AT  & 50.4& 44.6& 38.8& 28.0& 13.5& 35.1          \\
  \makecell{\textbf{PGD-AT}+\\ \textbf{ours}}     & \textbf{64.5} & \textbf{46.8} & \textbf{35.3} & \textbf{28.1} & \textbf{13.8} & \textbf{37.7} \\ \cdashline{1-7}[1pt/2pt]
  TRADES  & 42.1& 38.2& 34.0& 26.6& 15.1& 31.2          \\
  \makecell{\textbf{TRADES}+\\ \textbf{ours}}     & \textbf{62.5} & \textbf{47.9} & \textbf{39.2} & \textbf{26.7} & \textbf{15.1} & \textbf{38.3} \\ \cdashline{1-7}[1pt/2pt]
  MART & 41.0& 37.3& 33.6& 27.1& 16.4& 31.1          \\
  \makecell{\textbf{MART}+\\ \textbf{ours}} & \textbf{63.9} & \textbf{44.8} & \textbf{38.6} & \textbf{27.8} & \textbf{16.5} & \textbf{38.3} \\ \bottomrule
  \end{tabular}
  }\vspace{10pt}
  \caption{
    Adversarial accuracy on the tiny-ImageNet dataset.
  }
  \label{tab:tiny-imagenet}
\end{table}

\mypar{Performance on tiny-ImageNet.}
The comparison results on tiny-ImageNetNet are reported
in ~\cref{tab:tiny-imagenet}.
The results are consistent with the previous conclusion,
which shows a stable promotion of our method on a larger scale dataset.

\mypar{Other attacking methods.}
In this experiment, we compare we evaluate the adversarial accuracy under
other attacking methods, \eg PGD~\cite{madry2017towards}, FGSM ~\cite{goodfellow2014explaining}
and  C\&W~\cite{carlini2017towards}.
The attacking strength is fixed to $\epsilon = 8$.
Results are summarized in~\cref{tab:other_attack}.
Besides, we provide the performance of query-based attacking, \eg square attack~\cite{andriushchenko2020square},
in the supplementary material.

\begin{table}[!htb]
  \centering
  \resizebox{0.95\linewidth}{!}{
  \begin{tabular}{cccccc|cccccc}
  \toprule
  \multirow{2}{*}{Method} & \multicolumn{5}{c}{Attacking methods} &
  \multicolumn{5}{c}{Black-box attacking} \\
  \cline{2-12} 
     & std.   & PGD & FGSM& C\&W  & PGD+Ada.   & 0   & 1   & 2   & 4   & 8   & Avg \\
  \hline
  Standard   & 95.2 & 0.0 & 24.2& 0.0& -  & 95.2 & 0.0 & 24.2& 0.0& - \\
  \cdashline{1-6}[1pt/2pt]
  PGD-AT  & 86.9 & 46.2& 56.1& 45.3& -      & 86.9& 85.2& 83.1& 78.2& 66.8& 80.0    \\
  \makecell{\textbf{PGD-AT}+\\ \textbf{ours}}     & \textbf{95.8}  & \textbf{50.0} & \textbf{81.1} & \textbf{48.3} & 48.2 & \textbf{95.8} & \textbf{94.6} & \textbf{90.5} & \textbf{91.9} & \textbf{67.9} & \textbf{88.1}\\
  \cdashline{1-12}[1pt/2pt]
  TRADES  & 83.2 & 53.8& 64.6& \textbf{52.5}& -   & 83.2& 81.5   & 79.0& 74.1& 65.3& 76.6       \\
  \makecell{\textbf{TRADES}+\\ \textbf{ours}}      & \textbf{95.4}  & \textbf{52.6} & \textbf{80.3} & 51.0 & 52.2 & \textbf{95.4} & \textbf{94.7} & \textbf{92.7} & \textbf{82.4} & \textbf{65.9} & \textbf{86.2} \\
  \cdashline{1-12}[1pt/2pt]
  IAT  & 92.9 & 46.2& 65.7& 43.4& -  & 92.9& 91.2   & 88.7& 84.5& 71.8& 85.8 \\
  \textbf{IAT+ours} & \textbf{96.1} & \textbf{46.2}& \textbf{80.6}& \textbf{43.5}& 46.1 & \textbf{96.1} & \textbf{95.2} & \textbf{93.6} & \textbf{87.0} & \textbf{74.0} & \textbf{89.2} \\
  \cdashline{1-12}[1pt/2pt]
  MART & 83.6  & 55.3 & 64.3 & 53.4 & - & 83.6& 82.0& 80.0& 76.1& 66.0& 77.6 \\
  \makecell{\textbf{MART}+\\ \textbf{ours}}      & \textbf{95.9} & \textbf{56.0}& \textbf{81.3}& \textbf{53.6}& \textbf{55.6} & \textbf{95.9} & \textbf{95.3} & \textbf{93.4} & \textbf{83.7} & \textbf{67.7} & \textbf{87.2} \\ 
  \bottomrule
  \end{tabular}
  }
  \vspace{8pt}
  \caption{
    Left: adversarial accuracies under 3 different attacking methods and adaptive
    attack (PGD+Ada).
    Right: adversarial accuracies with black-box attacking.
    Our method is more robust against adversarial attacks compared to the adversarial training baselines.
    }
  \label{tab:other_attack}
\end{table}

The results in~\cref{tab:other_attack} tell that our method consistently outperforms the
competitors under various attacking methods and attacking strengths.
Specifically, under the FGSM attack, our method surpasses others with a significant
margin and the average performance improvement is nearly 15\%.

\mypar{Adaptive attack.}
Adaptive attacks~\cite{carlini2019evaluating} are specifically
crafted to compromise a proposed approach form a crucial component of adversarial
evaluation.
We test our method on the adaptive attack setting by exposing
the fusing weight, \ie $W_0$, to the attacker.
Concretely, we setup PGD
by combining the cross-entropy loss for classification outputs and
the binary-cross entropy (BCE) loss for weight generator
to jointly perform gradient ascent.
We test 7 different loss weights between CE and BCE
(10:1, 5:1, 2:1, 1:1, 1:2, 1:5, 1:10)
and report the best attacking results (worst adversarial accuracies) in
the last column of~\cref{tab:other_attack}.
More details about the settings of adaptive attack
are presented in the supplementary material.

The results of adaptive attacking are in~\cref{tab:other_attack}
(the `PGD+Ada' column).
The performance gap between PGD and PGD+Adv is marginal,
revealing that the proposed WG module is somehow robust against
adaptive attacking.

\mypar{Black-box robustness.}
Black-box attacks generate adversarial samples by attacking a surrogate model
which has a similar structure with the defense model and is trained on with similar tasks.
Following setting in ~\cite{wang2019improving,wang2019convergence},
we use ResNet-50 model as the surrogate model and it is trained on the CIFAR-10 dataset.
Then we generate adversarial samples by attacking the surrogate model and test the performance
of the defense model.

As shown in~\cref{tab:other_attack} (right), our method achieves better performance
compared to adversarial training baselines under black-box attacking.
This reveals that our method alleviates the `obfuscated gradients'
effect.
It can be verified by the following evidence~\cite{athalye2018obfuscated,zhang2019theoretically}:
(1) our method has higher accuracy under
weak attacks (\eg FGSM) than strong attacks (\eg PGD). (2) our method has higher
accuracy under black-box attacks than white-box attacks.

\subsection{Ablation Study} \label{sec:ablation}
In this section, we do several ablation experiments
on the CIFAR-10 dataset to verify\& interpret
our proposed framework.

\begin{figure}
  \begin{minipage}{.4\textwidth}
  \centering
  \includegraphics[width=\textwidth]{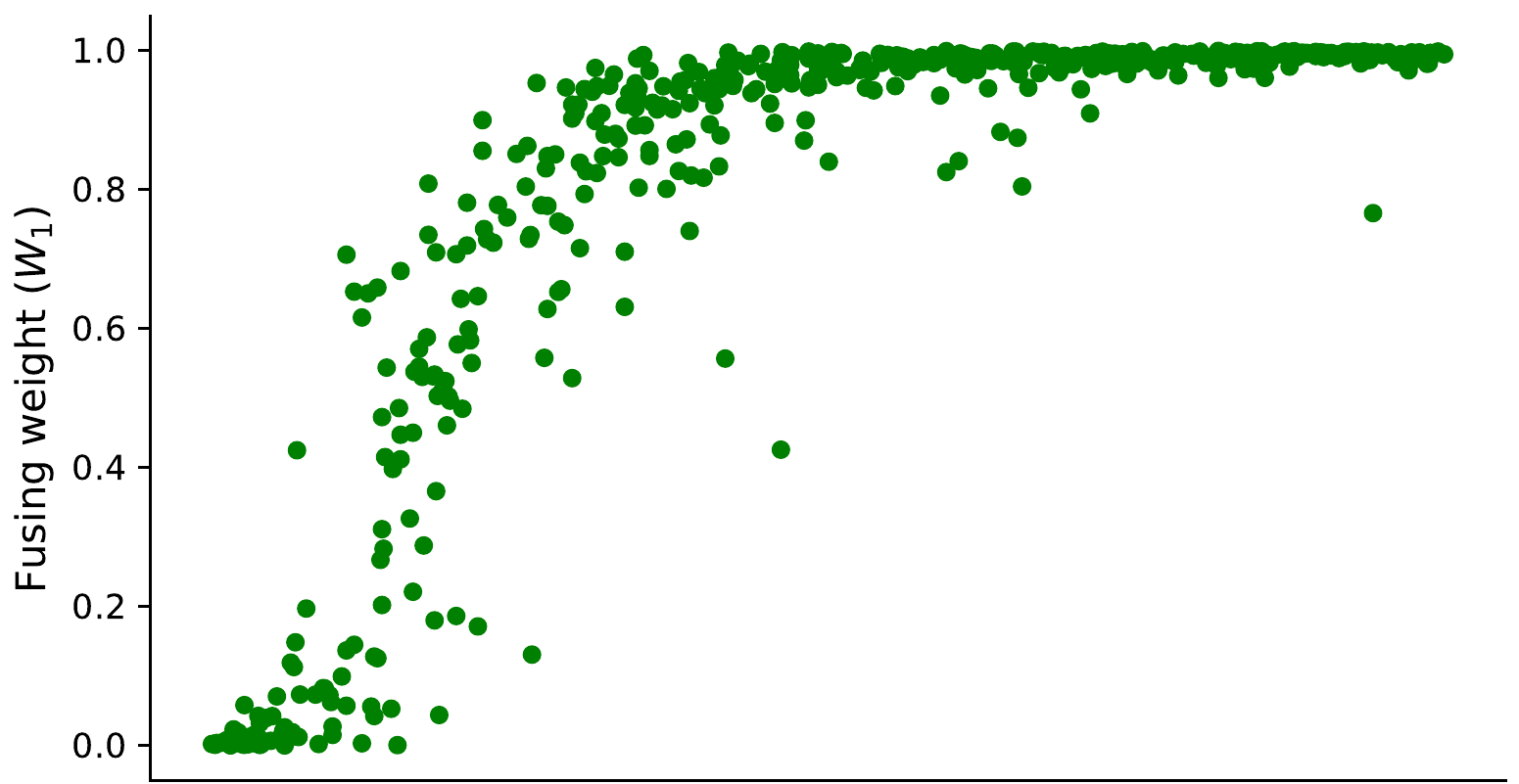}
  \captionof{figure}{Relationship between fusion weight $W_1$ and $\epsilon$.
  With the rising of $\epsilon$, $W_1$ increases, the features are biased
  towards the adversarial BN branch, \ie BN$_{k-1}$.}
  \label{fig:relationship}
  \end{minipage}\hfill
  \begin{minipage}{.58\textwidth}
      \centering
      \vspace{-10pt}
      \begin{tabular}{ccccccc}
        \toprule
        \multirow{2}{*}{\# BNs} & \multicolumn{6}{c}{Acc under different $\epsilon$} \\
        \cline{2-7} 
         & 0   & 1  & 2  & 4  & 8  & Avg      \\
         \hline
        $K=2$   & 95.0      & 85.1     & 82.2     & 70.4     & 48.1     & \bf{76.2}     \\
        $K=3$   & 95.5      & 86.7     & 83.2     & 72.3     & 48.6     & \bf{77.3}    \\
        $K=5$  & 95.8      & 87.0     & 83.9     & 72.6     & 50.0     & \bf{77.9}     \\
        $K=9$   & 95.6      & 87.2     & 84.5     & 72.9     & 50.4     & \bf{78.1}     \\
        \bottomrule
        \end{tabular}
  \captionof{table}{
    Ablation of the number of BN branches $K$ in stage I.
    In general, the performance improves with the raising of $K$.
  }
  \label{tab:different-k}
  \end{minipage}
\end{figure}

\mypar{Choice of $K$.}
Here we ablate the choice of $K$ when training the $K$-BN architecture in
stage I as illustrated in~\cref{fig:overall-architecture}.
Results in~\cref{tab:different-k} reveals that the model
benefits from a larger and the performance saturates at $K\ge5$.
For the compromise between performance and efficiency,
we use $K=5$ in our experiments.

\mypar{Relationship between fusion weight and $\epsilon$}
\label{sec:relationship_lambda_epsilon}
To further probe the behavior of AFA,
we analyze the relationship between fusion weights ($W_1$) and the attacking
strength ($\epsilon$).
Specifically, we train a ResNet-34 model on CIFAR-10
to generate adversarial samples of random attacking strengths.

As shown in~\cref{fig:relationship}, the AFA is biased towards the
the adversarial branch (BN$_{K-1}$) when the attacking
strength rises, as evidenced by the increase of $W_1$.
This nonlinear mapping is automatically learned by the weight generator.

\section{Conclusion}
In this paper, we proposed a simple yet effective framework for adversarial training.
We observed that the feature statistics of adversarial features transfer smoothly
with the change of attacking strength.
Based on the observation, we proposed an adaptive feature fusion framework to 
automatically align features for inputs of arbitrary attacking strength.
Compared to previous works, our method can handle input of different attacking
strengths on the fly with a single model, showing its potential in real-world applications.
We applied the proposed method to several recent adversarial training methods,
~\eg FGSM, PGD, TRADES, and MART, and tested the performance on SVHN, CIFAR-10, and tiny-ImageNet
datasets.
Extensive experiments demonstrated that our method improves the adversarial training
baselines with a considerable margin, under a wide range of attacking strengths.

\small{
  \bibliographystyle{unsrt}
  \bibliography{cvpr21defense.bib}
}
\end{document}